\documentclass[letterpaper, 10 pt, conference]{ieeeconf}

\IEEEoverridecommandlockouts 
\usepackage[T1]{fontenc}
\usepackage[utf8]{inputenc}
\usepackage{amsmath,amssymb} 
\usepackage{graphicx}
\usepackage{color}
\usepackage{subcaption}
\usepackage{multirow}
\usepackage{array,makecell,rotating}
\usepackage{hyperref}
\usepackage[normalem]{ulem}
\usepackage[table]{xcolor}
\usepackage{xspace}
\usepackage{soul}
\setstcolor{cyan}
\usepackage{caption}
\newcommand\nocell[1]{\multicolumn{#1}{c|}{}}

\newcommand{\gc}[0]{{\cellcolor{gray!15}}}

\usepackage{tikz}
\def\cmark{\tikz\fill[scale=0.4](0,.35) -- (.25,0) -- (1,.7) -- (.25,.15) -- cycle;}

\def\m#1{\mathbf #1}
\def\tightStar{\ensuremath\hspace{-0.22em}*\hspace{-0.12em}}

\xspaceaddexceptions{$)$}

\newcommand{\on}{\texttt{ON}\xspace}
\newcommand{\off}{\texttt{OFF}\xspace}
\newcommand{\unknown}{\texttt{UNKNOWN}\xspace}
\newcommand{\leftturn}{\texttt{LEFT TURN}\xspace}
\newcommand{\rightturn}{\texttt{RIGHT TURN}\xspace}
\newcommand{\flashers}{\texttt{FLASHERS}\xspace}
\newcommand{\vbehind}{\texttt{BEHIND}\xspace}
\newcommand{\vleft}{\texttt{LEFT}\xspace}
\newcommand{\vfront}{\texttt{FRONT}\xspace}
\newcommand{\vright}{\texttt{RIGHT}\xspace}

\graphicspath{{imgs/}}

\title{\LARGE \bf
        DeepSignals: Predicting Intent of Drivers Through Visual Signals
}

\author{
        Davi Frossard\textsuperscript{1,2} \qquad Eric Kee\textsuperscript{1} \qquad Raquel Urtasun\textsuperscript{1,2}
        \thanks{\textsuperscript{1}Uber Advanced Technologies Group}
        \thanks{\textsuperscript{2}University of Toronto}
        \thanks{\texttt{\footnotesize \{frossard, ekee, urtasun\}@uber.com}}
}

\begin{document}

\maketitle
\thispagestyle{empty}
\pagestyle{empty}

\begin{abstract}
Detecting the intention of drivers is an essential task in self-driving, necessary to anticipate sudden events like lane changes and stops. Turn signals and emergency flashers communicate such intentions, providing seconds of potentially critical reaction time. In this paper, we propose to detect these signals in video sequences by using a deep neural network that reasons about both spatial and temporal information. Our experiments on more than a million frames show high per-frame accuracy in very challenging scenarios. 
\end{abstract}

\section{Introduction}

Autonomous driving has risen as one of the most impactful applications of Artificial Intelligence (AI), where it has the  potential to change the way we live. 
Before self-driving cars are the norm however, humans and robots will have to share the roads. 
In this shared scenario, communications between vehicles are critical to alert others of  maneuvers that would otherwise be sudden or dangerous.
A social understanding of human intent is therefore essential to the progress of self-driving.
This poses additional complexity for self-driving systems, as such interactions are generally difficult to learn. 

Drivers communicate their intent to make unexpected maneuvers in order to give warning much further in advance than would otherwise be possible to infer from motion. Although driver movements communicate intent---for example when drivers slow down to indicate that they will allow a merge, or drive close to a lane boundary to indicate a desired merge position---motion cues are subtle, context dependent, and near-term. In contrast, visual signals, and in particular signal lights, are unambiguous and can be given far in advance to warn of unexpected maneuvers.

For example, without detecting a turn signal, a parked car may appear equally likely to remain parked as it is to pull into oncoming traffic. Analogously, when a driver plans to cut in front of another vehicle, they will generally signal in advance for safety. 
Buses also signal with flashers when making a stop to pick up and drop off passengers, allowing vehicles approaching from behind to change lanes, therefore reducing delays and congestion.

These everyday behaviors are safe when drivers understand the intentions of their peers, but are dangerous if visual signals are ignored.
Humans expect self-driving vehicles to respond.
We therefore consider in this work the problem of predicting driver intent through visual signals, and focus specifically on interpreting signal lights.

Estimating the state of turn signals  is a difficult problem: The visual evidence is small (typically only a few pixels), particularly at range, and occlusions are frequent. In addition, intra-class variations can be large. While some regulation exists, many vehicles have stylized blinkers, such as light bars with sequential lights in the direction being signaled, and the regulated frequency of blinking ($1.5 \pm 0.5$ Hz \cite{regulation}) is not always followed.
Furthermore, since we are interested in estimating intent, vehicle pose needs to be decoded. For instance, a left turn signal would correspond to a flashing light on the left side of a vehicle we are following, but on the other hand would correspond to a flashing light on the right side of an incoming vehicle. 
We refer the reader to \autoref{fig:challenge_ex} for an illustration of some of the challenges of turn signal estimation. 

\def\gapA{-1.3em}
\def\gapB{1em}
\begin{figure}[t]
	\centering
	\begin{tabular}{@{}c@{\hspace{0.5em}}c@{\hspace{0.5em}}c@{}}
        \includegraphics[width=0.3\columnwidth]{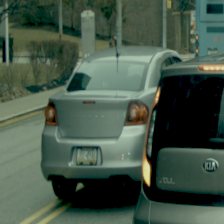} &
        \includegraphics[width=0.3\columnwidth]{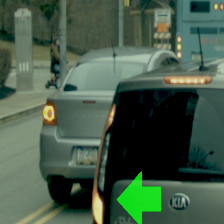} &
        \includegraphics[width=0.3\columnwidth]{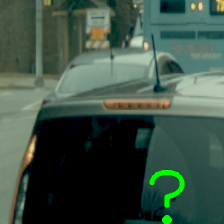} \\[0.25em]
        \includegraphics[width=0.3\columnwidth]{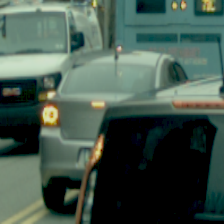} &
        \includegraphics[width=0.3\columnwidth]{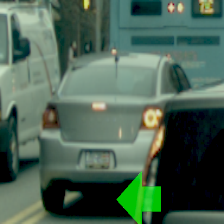} &
        \includegraphics[width=0.3\columnwidth]{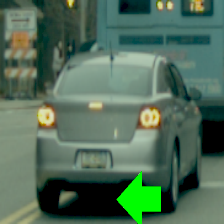}
	\end{tabular}
    \caption{A vehicle, signaling left, passes through occlusion. The actor's intent to turn left is correctly detected (left arrow), including the occlusion (question mark).}
    \label{fig:frontpage_example}
    \vspace{-2em}
\end{figure}

Surprisingly little work in the literature has considered this problem. Earlier published works \cite{frohlich2014will, casares2012robust} use hand-engineered features, trained in-part on synthetic data, and are evaluated on limited datasets. Other approaches have considered only nighttime scenarios \cite{7891988, satzoda2016looking}. Such methods are unlikely to generalize to the diversity of driving scenarios that are encountered every day.

In this paper, we identify visual signal detection as an important problem in self-driving. We introduce a large-scale dataset of vehicle signals, and propose a modern deep learning approach to directly estimate turn signal states from diverse, real-world video sequences. A principled network is designed to model the subproblems of turn signal detection: attention, scene understanding, and temporal signal detection. This results in a differentiable system that can be trained end-to-end using deep learning techniques, rather than relying upon hard coded premises of how turn signals should behave.

We demonstrate the effectiveness of our approach on a new, challenging real-world dataset comprising $34$ hours of video from our self-driving platform. The dataset includes the adverse conditions found in real-world urban driving scenarios, including occlusion, distant and uncommon vehicles, bad weather, and night/daytime exposures (see  \autoref{fig:challenge_ex} for an illustration). 

\def\gapA{-1.3em}
\def\gapB{0.7em}
\begin{figure}[t]
    \centering
    \begin{subfigure}[t]{0.3\columnwidth}
        \centering
        \includegraphics[width=\columnwidth]{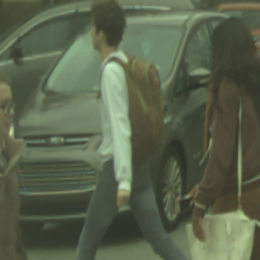}
        \vspace{\gapA}
        \caption{Occlusion}
        \vspace{\gapB}
    \end{subfigure}
    ~
    \begin{subfigure}[t]{0.3\columnwidth}
        \centering
        \includegraphics[width=\columnwidth]{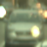}
        \vspace{\gapA}
        \caption{Distance}
        \vspace{\gapB}
    \end{subfigure}
    ~
    \begin{subfigure}[t]{0.3\columnwidth}
        \centering
        \includegraphics[width=\columnwidth]{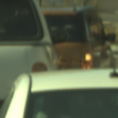}
        \vspace{\gapA}
        \caption{Queues}
        \vspace{\gapB}
    \end{subfigure}
    ~
    \begin{subfigure}[t]{0.3\columnwidth}
        \centering
        \includegraphics[width=\columnwidth]{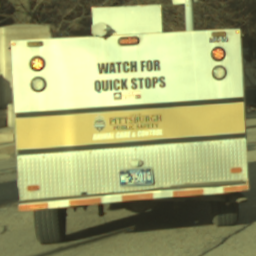}
        \vspace{\gapA}
        \caption{Unusual}
        \vspace{\gapB}
    \end{subfigure}
    ~
    \begin{subfigure}[t]{0.3\columnwidth}
        \centering
        \includegraphics[width=\columnwidth]{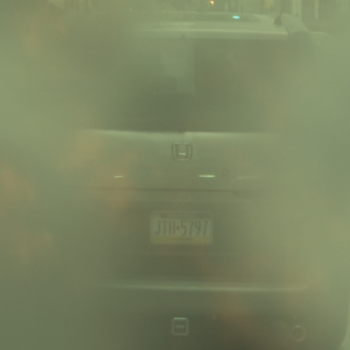}
        \vspace{\gapA}
        \caption{Weather}
        \vspace{\gapB}
    \end{subfigure}
    ~
    \begin{subfigure}[t]{0.3\columnwidth}
        \centering
        \includegraphics[width=\columnwidth]{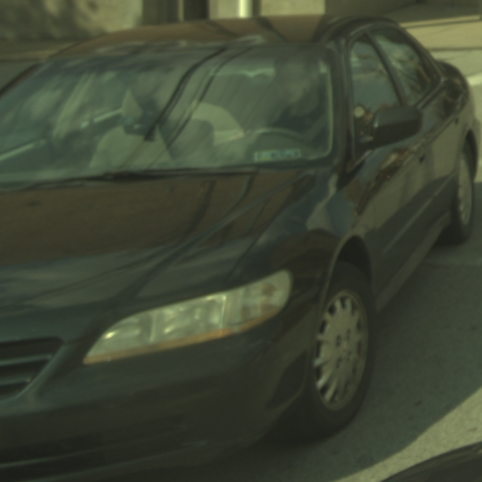}
        \vspace{\gapA}
        \caption{Bad Detections}
        \vspace{\gapB}
    \end{subfigure}
    ~
    \vspace{-1em}
    \caption{Challenging scenarios from the dataset of $1$,$257$,$591$ labeled frames.}
    \label{fig:challenge_ex}
    \vspace{-1em}
\end{figure}

\section{Related Work}

The field of visual perception for autonomous vehicles has been intensely explored in the past few years. Deep learning, and in particular deep convolutional neural networks (CNNs), has shown great improvements for tasks such as detection \cite{det1,det2,det3,det4}, tracking \cite{det0,det5}, instance and semantic segmentation \cite{seg1,seg2,seg3,seg4}, stereo estimation \cite{stereo1,stereo2,stereo3}, and (more related to the problem of action prediction) traffic understanding \cite{geiger20143d,scharwachter2014stixmantics,chen2015deepdriving}.

Little published work has been devoted to estimate the state of turn signals. Frohlich et al. \cite{frohlich2014will} proposed to detect light spots in vehicle bounding boxes and extract features using FFT over time, followed by an Adaboost classifier to predict the state of the turn signal. Another method \cite{casares2012robust} uses turn signals as a feature for vehicle detection and tracking. The state of the turn signals is also classified via a logic on the change of luminance on either side of the vehicle. Hsu et al. \cite{hsu2017learning} propose the use of SIFT flow to align a sequence of images as a pre-processing step to compute a difference image, which is then used by a CNN-LSTM architecture to predict both turn signal state as well as brake lights. These methods however use manually engineered features that do not adapt to different viewpoints of the vehicles (front, left, right), and testing considers a limited set of vehicles. 

Another line of work employs turn signals as a feature for vehicle detection. In \cite{7891988}, a Nakagami image is used to locate regions containing vehicular lights, which used as proposals for a Fast R-CNN detector. Other works use turn signals as a Haar-like feature for training Adaboost classifiers \cite{satzoda2016looking}.

In contrast, our work focuses on a fully learned approach to classify the state of turn signals using spatial and temporal features. This is closely related to recent works in the field of action recognition in videos, which has shown great progress with the use of deep learning. Among recent works, deep 3D convolutional neural networks for human action recognition in videos \cite{3DCNN} show the effectiveness of convolutions in the time dimension as a way to extract temporal features. Other works have considered decoupling spatial and temporal features \cite{simonyan2014two}, using two CNNs to perform action recognition: one processing single images for spatial features, and another using dense optical flow for temporal features.

Recurrent Neural Networks (RNNs) are an effective mechanism for temporal feature extraction. Recent works use LSTMs to learn social interactions between humans as a feature for tracking \cite{alahi2016social} and activity recognition \cite{DengVHM16} from videos. RNN-based agents have been trained with reinforcement learning in order to decide both how to tranverse the video and when to emit a prediction \cite{rnnagent}.

The union between CNNs and LSTMs has also been explored, with the use of CNNs to extract features from individual frames followed by LSTMs to incorporate temporal features \cite{cnnlstm,cnnlstm2}. This approach can be extended by using 3D CNNs to extract local spatio-temporal features and a LSTM for long term temporal features \cite{3dcnnlstm}. Going a step further, CNNs and LSTMs can be fused into a single architecture by using convolutional gates, an approach first introduced in order to predict precipitation from radar intensity images \cite{convlstm}, but that has been recently used to create online maps for self driving vehicles \cite{Homayounfar_2018_CVPR}.

While initially introduced for machine translation \cite{bahdanau2014neural}, attention mechanisms have also been for neural captioning \cite{xu2015show} and action recognition \cite{sharma2015action}. In \cite{rnnevents}, a Bidirectional LSTM is used to both detect events in videos and attend to the agents causing the event. ConvLSTMs with attention have also been used for action classification \cite{li2018videolstm}.

\section{Predicting Actor Intent Through Visual Signals}

\begin{figure*}[t!] 
    \includegraphics[width=\textwidth]{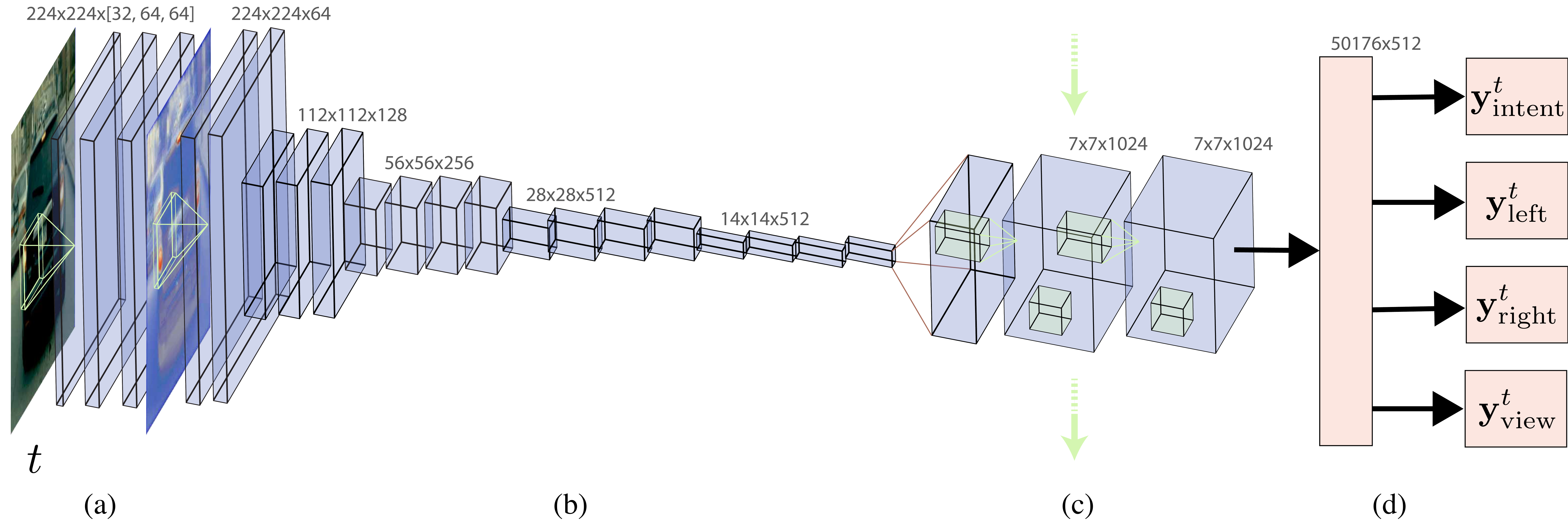}
    \caption{In this work, we propose the use of a convolutional-recurrent architecture in order to classify the state of turn signals in vehicles. For each input frame, an attention mask is predicted using a fully convolutional network (a), the element-wise product is then taken with the original input image and spatial features are extracted using a VGG16-based CNN (b), temporal features are then incorporated using a Convolutional LSTM (c). From the final hidden state, probability distributions are predicted over the state of turn signals and the view face of the vehicle (d).}
    \label{fig:tsnet}
    \vspace{-1em}
\end{figure*}

In this paper, we tackle the problem of estimating actor intent from the state of turn signals and emergency lights. These visual signals are naturally captured in  
video on autonomous driving platforms.
 A typical video scene can contain multiple vehicles, which presents the problem of detection and tracking. Here it is assumed that such methods are available, and a preprocessing step will be applied to recover an axis-aligned region of interest (ROI) around each actor. 

The ROI around an actor constitutes a streaming input of cropped video frames. The desired system output is also a stream, indicating the state of the actor's signals at each frame. Classifying turn signal states is a challenging problem,
as the left turn signal of a vehicle can appear on either the image's left or right depending upon the vehicle orientation. The network therefore must also resolve the pose of the actor to correctly classify turn signal state regardless of view face.

Within any particular video frame, a signal light may be illuminated or not, while its logical state is \on, \off, or \unknown (the latter representing when the light is occluded). Given the logical state of both signal lights, the logical state of the turn signal can be determined: \leftturn, \rightturn, \flashers, \off, or \unknown. For example, observing $($\on, \off$)$ for the left and right signal lights indicates a \leftturn. Note that $($\on, \unknown$)$ can be mapped to either \leftturn or \unknown depending upon how much confidence is desired. In this section, we describe a learned model to predict such turn signal states.

\subsection{Model formulation}

The model is formulated to address three subproblems: {\it  attention,} to identify the signal lights of the actor; {\it semantic understanding}, to identify occlusion and the direction from which the actor is being viewed; and {\it  temporal reasoning}, to distinguish flashing lights and persistent lights from other specious patterns. A deep learning architecture is designed to address these problems. We refer the reader to \autoref{fig:tsnet} for an illustration. Input frames are first processed by an attention module, which applies a spatial mask, and a deep convolutional network is used to recover spatial concepts. This per-frame information is then input to a convolutional LSTM to distinguish the temporal patterns of turn signals and emergency flashers from other content. The resulting spatial and temporal features are passed to fully connected layers for classification.

Attention processing begins by resizing the input ROI to a fixed $224\times 224$ pixels. A \mbox{$4$-layer}, fully convolutional network is used to compute a pixel-wise, scalar attention value. Kernels are $3\times 3$ with dilations $(1, 2, 2, 1)$ and channel dimensions $(32, 64, 64, 1)$. The resulting scalar mask is point-wise multiplied with the original, resized input ROI. This allows the network to both add more saliency to relevant pixels and attenuate noisy spatial artifacts.

Spatial features are extracted using a  CNN architecture based on VGG16 \cite{vgg},  as shown in \mbox{\autoref{fig:tsnet} (b)}. Weights are pre-trained on ImageNet, and fine tuned during training. This allows the network to model the vehicle of interest, its orientation, occluding objects, and other spatial concepts. The $7\times 7\times 512$ output retains a coarse spatial dimension for temporal processing by a Convolutional LSTM.

A Convolutional LSTM (ConvLSTM) module \cite{convlstm} is used to refine the spatial features by modeling temporal characteristics of the streaming input (now feature tensors), depicted in \mbox{\autoref{fig:tsnet} (c)}. Note that this design factors the spatial and temporal components of the model into separate modules. We show that factorization more efficiently uses the available compute resources, and leads to greater performance. The ConvLSTM allows the network to reason through time, and thus distinguish between flashing and persistent lights (i.e., turn signals and brake lights).

Convolutional LSTMs learn temporal representations by maintaining an internal, hidden state, which is modified through a series of control gates. Let $\m X_t$ be the feature tensor that is input at time $t$, and $\m W$ and $\m B$ be the learned weights and biases of the ConvLSTM. The hidden state is embodied by two tensors, $\m H$ and $\m C$, which are updated over time by the following expressions:
\def\neg{\text{-}}
\begin{align}\label{eq:input}
    \m I_t & = \sigma\big(\m W^{xi} \tightStar \m X_t + \m W^{hi} \tightStar \m H_{t\neg 1} + \m W^{ci} \tightStar \m C_{t \neg 1} + \m B^{i} \big)
\\[0.5em]\label{eq:forget}
    \m F_t
    & = \sigma\big( \m W^{xf} \tightStar \m X_t + \m W^{hf} \tightStar \m H_{t\neg 1} + \m W^{cf} \tightStar \m C_{t\neg 1} + \m B^{f} \big)
\\[0.5em]\nonumber
    \m C_t 
    & = \m F_t \circ \m C_{t\neg 1}	+	\dots
\\\label{eq:cell}
	& \hspace{2.5em}
	\m I_t	\circ	\tanh\big( \m W^{xc} \tightStar \m X_t + \m W^{hc} \tightStar \m H_{t\neg 1} + \m B^{c}  \big)
\\[0.5em]\label{eq:output}
    \m O_t 
    & = \sigma\big( \m W^{xo} \tightStar \m X_t + \m W^{ho} \tightStar \m H_{t\neg 1} + \m W^{co} \tightStar \m C_{t} + \m B^{o} \big)
\\[0.5em]\label{eq:hidden}
    \m H_t
    & = \m O_t \circ	\tanh\big( \m C_t	\big)~~.
\end{align}

The parameterized gates $\m I$ (input), $\m F$ (forget) and $\m O$ (output) control the flow of information through the network, and how much of it should be propagated in time. Temporal information is maintained through the cell memory, which accumulates relevant latent representations, as shown in Equation~(\ref{eq:cell}). Note that, to prevent overfitting, we apply dropout on the output of Equation~(\ref{eq:cell}) as a regularizer. Specifically, the input gate controls the use of new information from the input, Equation~(\ref{eq:input}); the forget gate controls what information is discarded from the previous cell state, Equation~(\ref{eq:forget}); and the output gate further controls the propagation of information from the current cell state to the output, Equation~(\ref{eq:output}) by element-wise multiplication, Equation~(\ref{eq:hidden}). 

The ConvLSTM module is constructed as a series of ConvLSTM layers, each following Equations~(\ref{eq:input})-(\ref{eq:hidden}). In this multi-layer architecture, each subsequent layer takes as input the hidden state, $\m H_t$, from the preceding layer (the first layer takes $\m X_t$ as input). In particular, we use two ConvLSTM layers, each with a $7\times 7\times 256$ hidden state.

Lastly, the ConvLSTM features are passed through a fully connected layer, depicted in  \autoref{fig:tsnet} (d). This produces the random variables of interest: $\m y^t_{\text{intent}}$ over the states \leftturn, \rightturn, \flashers, \off and \unknown; $\m y^t_{\text{left}}$ and $\m y^t_{\text{right}}$ over the states \on, \off, \unknown (for the individual lights on the left and right sides of the vehicle, respectively). We also define states from which an actor is being viewed from \vbehind, \vleft, \vfront, \vright, and show that predicting these as $\m y^t_{\text{view}}$ helps learning. 

\subsection{Learning}
We train our  model  using a multi-task loss. Specifically, a weighted cross entropy loss is defined over the tasks. Given model inputs $x$, ground truth labels $\hat{y}$, model weights $\theta$, task weights $\gamma$ and the network function $\sigma(\cdot)$, the loss is
\begin{align}\label{eq:loss_fun}
    \notag\mathcal{L}(\hat{y},x|\theta) &= \ell_{intent}(\hat{y},x|\theta) \\
    \notag                              &+ \ell_{left}(\hat{y},x|\theta) + \ell_{right}(\hat{y},x|\theta) \\
                                        &+ \ell_{view}(\hat{y},x|\theta) 
\end{align}
where each task loss uses cross-entropy   defined as 
\begin{equation*}
    \ell(\hat{y},x|\theta) = \gamma \sum_c \hat{y}_c \log(\sigma_c(x|\theta))
\end{equation*}

Note that the loss is defined in terms of a sum over the task space, which includes: $\ell_{intent}$, the loss over the high level understanding of the actor; $\ell_{left}$ and $\ell_{right}$, the losses over the left and right turn signals, respectively; and $\ell_{view}$, the loss over the face of the actor that is seen. We also use L2 weight decay on dense layers to prevent overfitting.

\section{Experimental Evaluation}

\begin{figure}[t]
    \centering
    \begin{subfigure}[t]{0.48\columnwidth}
        \centering
        \includegraphics[trim={0.15in, 0.37in, 0.15in, 0.13in}, clip, width=\columnwidth]{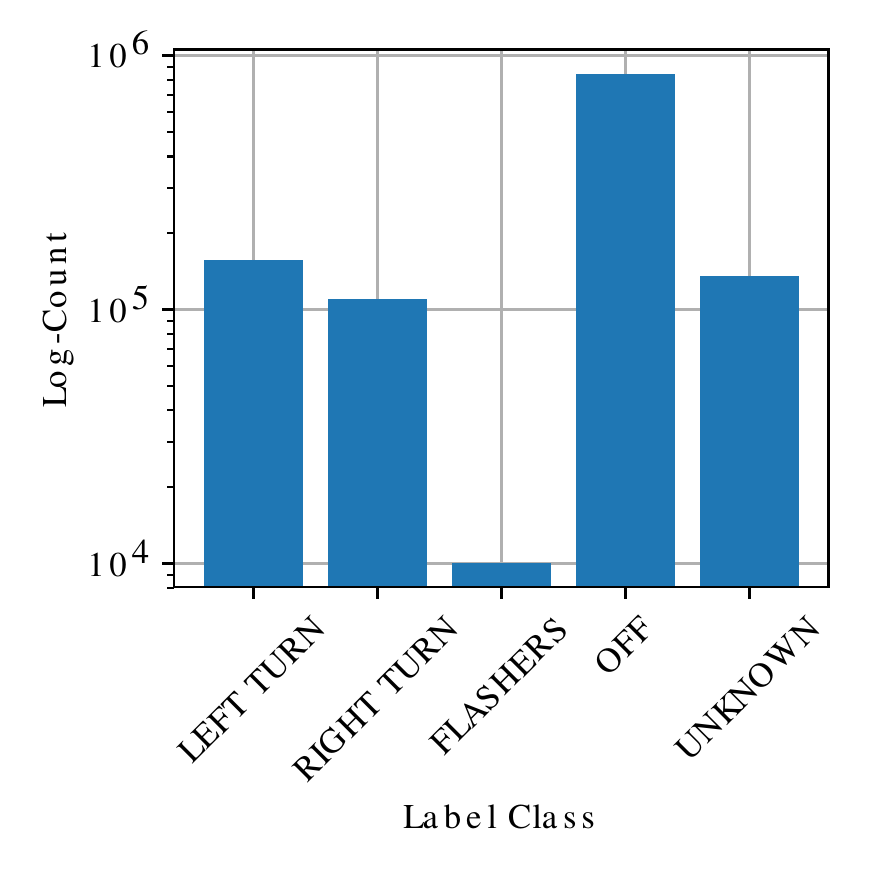}
        \caption{Label classes} 
        \label{tab:dual_counts}
    \end{subfigure}
    \begin{subfigure}[t]{0.48\columnwidth}
        \centering
        \includegraphics[trim={0.15in, 0.37in, 0.15in, 0.13in}, clip, width=\textwidth]{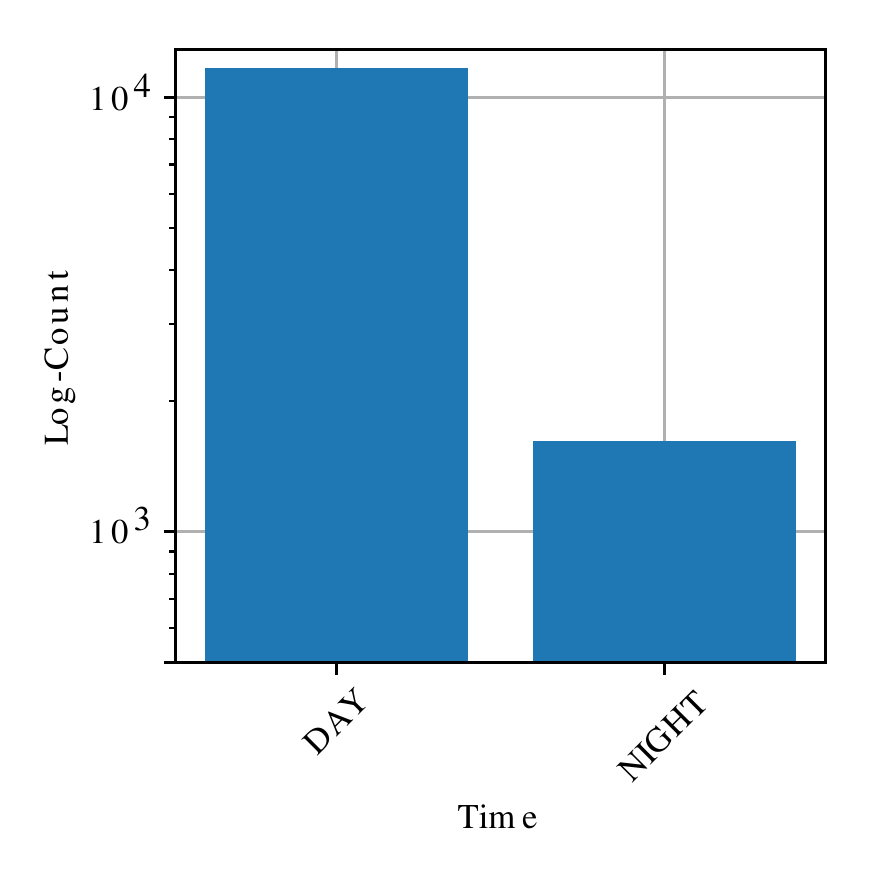}
        \caption{Label time} 
        \label{tab:view_counts}
    \end{subfigure}
    \begin{subfigure}[t]{0.48\columnwidth}
        \centering
        \includegraphics[trim={0.15in, 0.37in, 0.15in, 0.13in}, clip, width=\columnwidth]{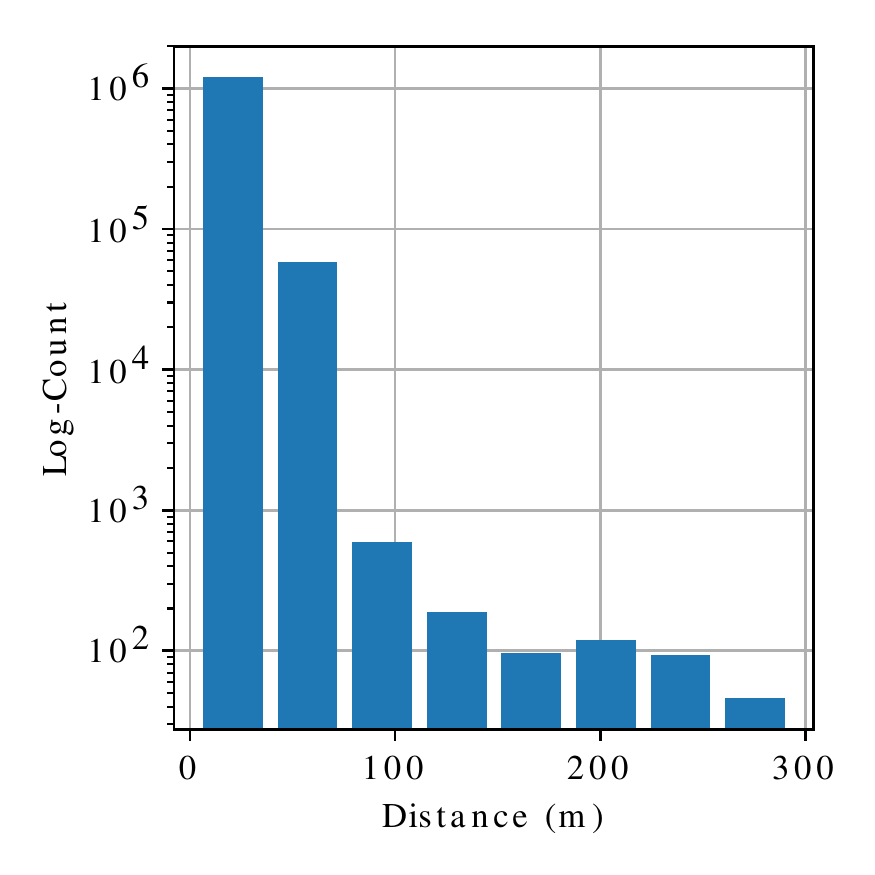}
        \caption{Label distances (m)} 
        \label{tab:dist_counts}
    \end{subfigure}
    \begin{subfigure}[t]{0.48\columnwidth}
        \centering
        \includegraphics[trim={0.15in, 0.37in, 0.15in, 0.13in}, clip, width=\textwidth]{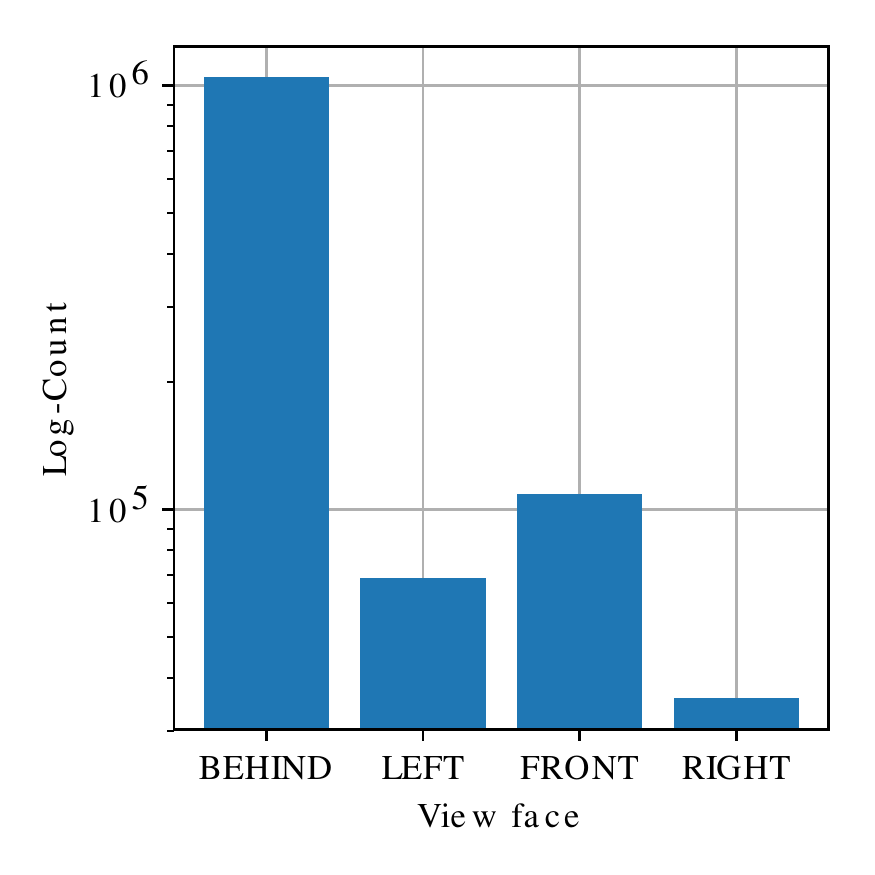}
        \caption{Label view faces} 
        \label{tab:view_counts}
    \end{subfigure}
    \caption{Dataset distributions in log units. The distribution over labels (a) is imbalanced towards \texttt{OFF}, with \texttt{FLASHERS} being least common. Number of sequences during daytime (b) is almost an order of magnitude larger than those during nighttime.  Distance to vehicles (c) place 90\% within 30 meters. Lastly, the majority of vehicles are seen from behind (d), with right being relatively under-represented.} 
    \label{fig:histograms}
\end{figure}

\paragraph{Dataset} Unfortunately, there is no public dataset for turn signal classification. Therefore, we label over 10,000 vehicle trajectories recorded in our autonomous driving platform at 10 Hz in terms of the state of turn signals, for a total of 1,257,591 labeled frames. Each frame is labeled for the left turn and right turn lights in terms of \texttt{ON}, \texttt{OFF} or \texttt{UNKNOWN}. Note that the label identifies the conceptual state of each light, with \texttt{ON} indicating that the signal is active even when the light bulb is not illuminated. These lower level labels are used to infer the high level action intents: \texttt{LEFT TURN}, \texttt{RIGHT TURN}, \texttt{FLASHERS}, \texttt{OFF} and \texttt{UNKNOWN}, which ultimately is what the model is trained to classify. The number of labels is shown in \autoref{tab:dual_counts} and it evidences a considerable bias towards the \texttt{OFF} class in the dataset. Also shown are the distributions over distance (\autoref{tab:dist_counts}) and viewpoint (\autoref{tab:view_counts}).

\paragraph{Experimental Setup} To train the models, we use Adam optimization \cite{adam} with a learning rate of $1\times10^{-4}$, $\beta_1 = 0.9$, and $\beta_2 = 0.999$. We also reduce the learning rate on plateau, multiplying it by a factor of $0.1$ if 5 epochs go by without changing the loss by more than $1\times10^{-3}$. A weight decay of $1\times10^{-4}$ and dropout with $p=0.5$ is used in the fully connected layers for regularization. Training mini-batches are sampled using a stratified scheme that counteracts class imbalance. Training is limited to 50 epochs (models will generally converge around the 25th epoch) and selection is done according to the validation metrics. Lastly, data augmentation via random mirroring and color jittering is applied to the input sequences.

\begin{table}[t]
    \centering
    \setlength{\tabcolsep}{0.5em}
    \def\arraystretch{1.2}
    \begin{tabular}{|c|c|c|c|c|c|c|}
        \hline
        \textbf{\textsc{Model}} & \textbf{\textsc{Accuracy}} & \textbf{\textsc{Recall}} & \textbf{\textsc{F1}} & \textbf{\textsc{Fp}} & \textbf{\textsc{Fn}}\\\hline
        FC-LSTM         & 35.30\%          & 30.47\%          & 32.71\%          & 27.70\%           & 61.65\%          \\\hline
        ConvLSTM        & 37.32\%          & 34.42\%          & 35.81\%          & 30.07\%           & 63.95\%          \\\hline
        CNN-LSTM        & 60.52\%          & 60.54\%          & 60.53\%          & 11.21\%           & 39.17\%          \\\hline
        \textbf{ours}   & \textbf{70.89\%} & \textbf{72.11\%} & \textbf{71.49\%} &  \textbf{5.63\%}  & \textbf{24.00\%} \\\hline
    \end{tabular}
    \caption{Comparison to baseline systems.}
    \label{tab:baseline_results}
    \vspace{-1em}
\end{table}

\begin{table}[t]
\centering
\setlength{\tabcolsep}{0.15em}
\def\arraystretch{1.2}
\resizebox{\columnwidth}{!}{%
\begin{tabular}{|>{\centering\arraybackslash}p{1.7cm}|>{\centering\arraybackslash}p{1.4cm}|>{\centering\arraybackslash}p{1.4cm}|>{\centering\arraybackslash}p{1.4cm}|>{\centering\arraybackslash}p{1.4cm}|>{\centering\arraybackslash}p{1.4cm}|}
    \cline{2-6}
    \nocell{1} & \textbf{\textsc{Left}} & \textbf{\textsc{Right}} & \textbf{\textsc{Flashers}} & \textbf{\textsc{Off}} & \textbf{\textsc{Unknown}} \\ \hline
    \textbf{\textsc{Left}}  & \gc74\%  &  1\%     &  0\%     & 22\%    &  2\%     \\\hline
    \textbf{\textsc{Right}} &  1\%     & \gc77\%  &  0\%     & 18\%    &  4\%     \\\hline
    \textbf{\textsc{Emergency}}  &  9\%     & 17\%     & \gc36\%  & 26\%    & 13\%    \\\hline
    \textbf{\textsc{Off}}        &  3\%     &  2\%     &  0\%     & \gc92\% &  3\%     \\\hline
    \textbf{\textsc{Unknown}}    &  3\%     &  1\%     &  1\%     & 23\%    & \gc72\% \\\hline
    \end{tabular}
}
    \caption{Confusion matrix for our proposed model.} 
    \label{tab:conv_lstm_results} 
	\vspace{-2em}
\end{table}

\paragraph{Baselines} We evaluate our proposed method against a series of baselines. In particular, it is compared against a fully connected LSTM (FC-LSTM), a Convolutional LSTM (ConvLSTM), and an LSTM using convolutional features (CNN-LSTM). In all cases, we use sequences of vehicle observations at size $224 \times 224$ pixels. For the FC-LSTM, the sequences are flattened and passed through an FC-LSTM with 3 hidden layers (with 256, 256 and 512 neurons, respectively). For the ConvLSTM, we also use a 3 layer network (with 8, 8 and 3 channels, respectively (deeper models could not be fit in the GPU). Lastly, for the CNN-LSTM, features are first extracted using VGG16, flattened and fed to a two layer LSTM with 256 and 128 neurons.

\paragraph{Metrics} To evaluate the model, we use accuracy, recall, and F1 metrics. We also define False Positives (FP) as being the cases in which a ground-truth \texttt{OFF} or \texttt{UNKNOWN} is classified as any other state, and False Negatives (FN) when a ground-truth \texttt{ON} is classified as any other state.

The aforementioned metrics are shown in \autoref{tab:baseline_results} for the FC-LSTM, ConvLSTM, CNN-CLSTM and our proposed method. The FC-LSTM results in the weakest performance; this can be explained by the ineffectiveness of fully connected layers in extracting spatial features, counterbalanced only by the large capacity of the network, which allows it to learn more complex functions. The ConvLSTM achieves slightly better results by leveraging convolutions in the gates, which makes it more suitable for spatial feature extraction. Its memory inefficiency, however, prevents us from using deeper architectures and therefore limits the capacity of the model. Combining the two  approaches we arrive at a CNN-LSTM, which is both able to leverage the rich spatial feature extraction from CNNs and temporal feature representation of LSTMs, achieving better results than the previous baselines. Our proposed method further adds convolutions inside the LSTM and attention mechanisms, giving the best results.

\begin{figure*}[t]
    \centering
    \begin{subfigure}[t]{0.2\textwidth}
        \centering
        \includegraphics[width=\textwidth]{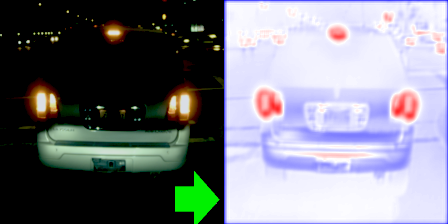}
    \end{subfigure}
    ~~
    \begin{subfigure}[t]{0.2\textwidth}
        \centering
        \includegraphics[width=\textwidth]{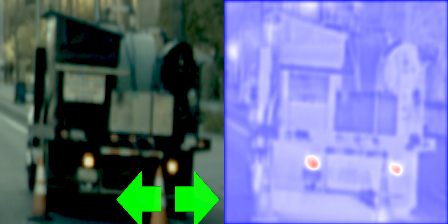}
    \end{subfigure}
    ~~
    \begin{subfigure}[t]{0.2\textwidth}
        \centering
        \includegraphics[width=\textwidth]{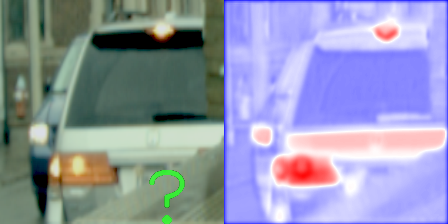}
    \end{subfigure}
    ~~
    \begin{subfigure}[t]{0.2\textwidth}
        \centering
        \includegraphics[width=\textwidth]{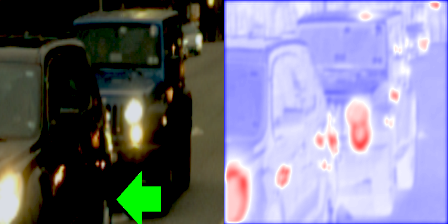}
    \end{subfigure}
    \\
    \begin{subfigure}[t]{0.2\textwidth}
        \centering
        \includegraphics[width=\textwidth]{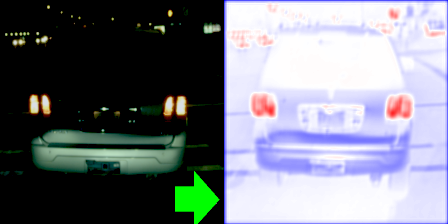}
    \end{subfigure}
    ~~
    \begin{subfigure}[t]{0.2\textwidth}
        \centering
        \includegraphics[width=\textwidth]{more_viz/2/1.png}
    \end{subfigure}
    ~~
    \begin{subfigure}[t]{0.2\textwidth}
        \centering
        \includegraphics[width=\textwidth]{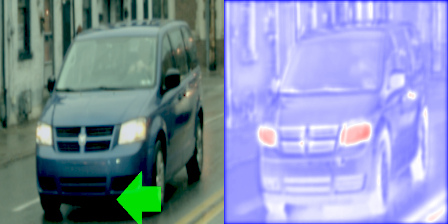}
    \end{subfigure}
    ~~
    \begin{subfigure}[t]{0.2\textwidth}
        \centering
        \includegraphics[width=\textwidth]{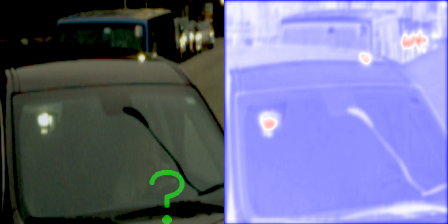}
    \end{subfigure}
    \\
    \begin{subfigure}[t]{0.2\textwidth}
        \centering
        \includegraphics[width=\textwidth]{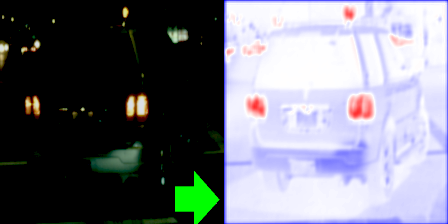}
    \end{subfigure}
    ~~
    \begin{subfigure}[t]{0.2\textwidth}
        \centering
        \includegraphics[width=\textwidth]{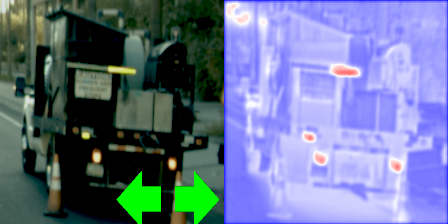}
    \end{subfigure}
    ~~
    \begin{subfigure}[t]{0.2\textwidth}
        \centering
        \includegraphics[width=\textwidth]{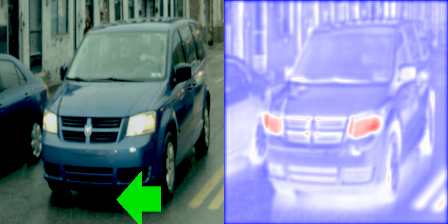}
    \end{subfigure}
    ~~
    \begin{subfigure}[t]{0.2\textwidth}
        \centering
        \includegraphics[width=\textwidth]{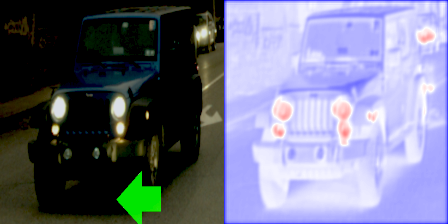}
    \end{subfigure}
    \\
    \begin{subfigure}[t]{0.2\textwidth}
        \centering
        \includegraphics[width=\textwidth]{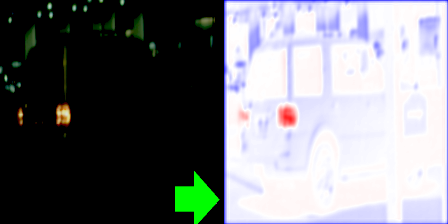}
        \caption{}
    \end{subfigure}
    ~~
    \begin{subfigure}[t]{0.2\textwidth}
        \centering
        \includegraphics[width=\textwidth]{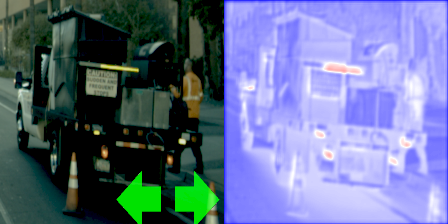}
        \caption{}
    \end{subfigure}
    ~~
    \begin{subfigure}[t]{0.2\textwidth}
        \centering
        \includegraphics[width=\textwidth]{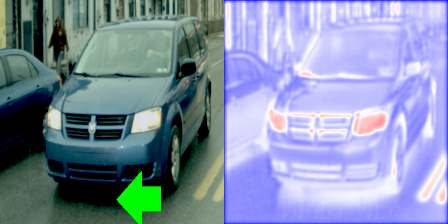}
        \caption{}
    \end{subfigure}
    ~~
    \begin{subfigure}[t]{0.2\textwidth}
        \centering
        \includegraphics[width=\textwidth]{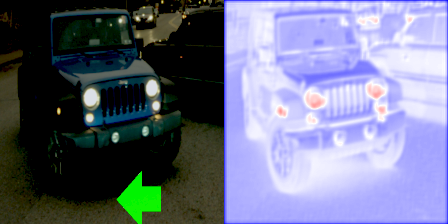}
        \caption{}
    \end{subfigure}
    \caption{Several sequences from the test set. Input image is shown on the left with the network output illustrated on the bottom right, the corresponding attention mask is shown on the right. Column (a), right turn signal is correct before and throughout the maneuver. Columns (b), vehicle with flashers stopped on the sidewalk is correctly classified. Columns (c) and (d) show challenging sequences with correct classification of incoming vehicles signaling left turns (including occlusion).}
    \label{fig:net_examples_various}
\end{figure*}

\begin{figure*}[t!] 
    \newlength{\imagewidth}
    \settowidth{\imagewidth}{\includegraphics{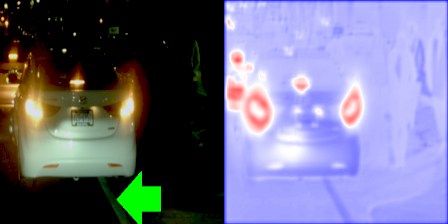}}
    \centering
    \begin{subfigure}[t]{0.3\columnwidth}
        \centering
        \includegraphics[trim=0 0 0.5\imagewidth{} 0,clip,width=\columnwidth]{fail_modes/0.png}
        \caption{~}
    \end{subfigure}
    ~
    \begin{subfigure}[t]{0.3\columnwidth}
        \centering
        \includegraphics[trim=0 0 0.5\imagewidth{} 0,clip,width=\columnwidth]{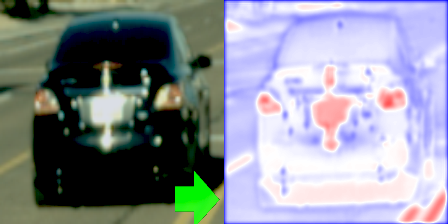}
        \caption{~}
    \end{subfigure}
    ~
    \begin{subfigure}[t]{0.3\columnwidth}
        \centering
        \includegraphics[trim=0 0 0.5\imagewidth{} 0,clip,width=\columnwidth]{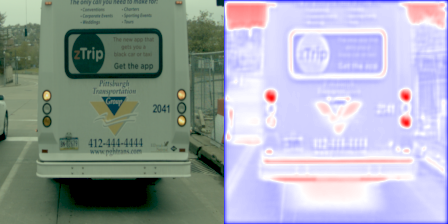}
        \caption{~}
    \end{subfigure}
    ~
    \begin{subfigure}[t]{0.3\columnwidth}
        \centering
        \includegraphics[trim=0 0 0.5\imagewidth{} 0,clip,width=\columnwidth]{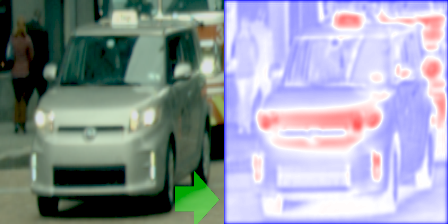}
        \caption{~}
    \end{subfigure}
    ~
    \begin{subfigure}[t]{0.3\columnwidth}
        \centering
        \includegraphics[trim=0 0 0.5\imagewidth{} 0,clip,width=\columnwidth]{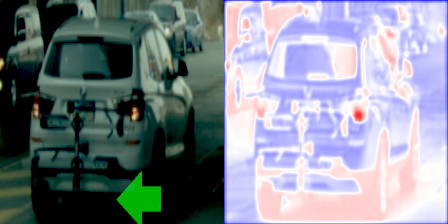}
        \caption{~}
    \end{subfigure}
    \caption{Failure modes of the network. (a) Bright lights at night are misclassified as a left turn. (b) Bright reflection on the right side of a distant vehicle is misclassified as a right turn. (c) Right turn signal is missed on an unusual vehicle. (d) Actor pose is incorrectly decoded and output is flipped. (e) False positive left turn on a vehicle carrying a bike.}
    \label{fig:failure_modes}
\end{figure*}

The confusion matrix of our approach is shown in \autoref{tab:conv_lstm_results}. 
Note that detection accuracy is distinctly high for \texttt{OFF} and low for \texttt{FLASHERS} because the two classes are over- and under-represented (respectively) in the dataset, \autoref{tab:view_counts} (a). This imbalance naturally affects test performance, even when using a stratified sampling scheme, because the label distributions are similar between train and test sets.

Lastly, we show several outputs from our model in \autoref{fig:frontpage_example} and \autoref{fig:net_examples_various}. These examples illustrate that turn signals can be classified from various viewpoints, and that occlusion can be detected. Furthermore, the attention network produces masks with highest value around bright spots, which makes turn signal regions more salient, improving performance (see Section~\ref{sec:ablation}). In \autoref{fig:failure_modes}, we show examples of frames in which the network fails, and provide hypotheses for the cause. Failures can result from distracting lights (or reflections), and unusual vehicle shapes or features. Examples are also included as video in the supplementary material.

\subsection{Ablation Studies}
\label{sec:ablation}

Here we evaluate the effects of diverse changes to the model: using real detections, changing the attention architectures, output parameterization and normalization schemes. Results for the experiments are shown in Tables \ref{tab:detections}-\ref{tab:studies}.

\begin{table}[t]
    \centering
    \setlength{\tabcolsep}{0.5em}
    \def\arraystretch{1.2}
    \begin{tabular}{|c|c|c|c|c|c|c|}
        \hline
        \textbf{\textsc{Input}} & \textbf{\textsc{Accuracy}} & \textbf{\textsc{Recall}} & \textbf{\textsc{F1}} & \textbf{\textsc{Fp}} & \textbf{\textsc{Fn}}\\\hline
        Detections & 66.85\%          & 62.96\%          & 64.85\%          &  7.42\%           & 26.25\%          \\\hline
        Labels     & \textbf{70.89\%} & \textbf{72.11\%} & \textbf{71.49\%} &  \textbf{5.63\%}  & \textbf{24.00\%} \\\hline
    \end{tabular}
    \caption{Comparison between labels and detections as input for the model (numbers reported on held-out test set).}
    \vspace{-1.5em}
    \label{tab:detections}
\end{table}

\begin{table*}[t]
    \centering
    \setlength{\tabcolsep}{0.3em}
    \def\arraystretch{1.4}
    \begin{tabular}{rc|c|c|c|c|c|c|c|c|c|}
        \cline{3-11}
            & & \multirow{2}{*}{\textbf{\textsc{Attention}}} & \multicolumn{3}{c|}{\textbf{\textsc{Loss}}}                  
            & \multirow{2}{*}{\textbf{BN}} & \multirow{2}{*}{\textbf{LN}} & \multirow{2}{*}{\textbf{\textsc{Accuracy}}} & \multirow{2}{*}{\textbf{\textsc{Recall}}} & \multirow{2}{*}{\textbf{\textsc{F1}}} \\ \cline{4-6}
            &                   &                                         & $\ell_{intent}$ & $\ell_{left}/\ell_{right}$ & $\ell_{view}$ &                              &                              &                                         &                                           &                   \\ \cline{3-11} 
            & {\scriptsize $1$} & ---                                     & \cmark          & \cmark                     & \cmark        & ---                          & ---                          & 66.03\%                                 & 66.63\%                                   & 66.33\%           \\ \cline{3-11} 
            & {\scriptsize $2$} & \cmark$\tightStar$                      & \cmark          & \cmark                     & \cmark        & ---                          & ---                          & 66.82\%                                 & 70.84\%                                   & 68.77\%           \\ \cline{3-11} 
            & {\scriptsize $3$} & \cmark                                  & \cmark          & ---                        & ---           & ---                          & ---                          & 63.77\%                                 & 68.95\%                                   & 66.26\%           \\ \cline{3-11} 
            & {\scriptsize $4$} & \cmark                                  & \cmark          & ---                        & \cmark        & ---                          & ---                          & 65.60\%                                 & 70.89\%                                   & 68.14\%           \\ \cline{3-11} 
            & {\scriptsize $5$} & \cmark                                  & \cmark          & \cmark                     & \cmark        & \cmark                       & ---                          & 61.01\%                                 & 67.29\%                                   & 65.61\%           \\ \cline{3-11} 
            & {\scriptsize $6$} & \cmark                                  & \cmark          & \cmark                     & \cmark        & ---                          & \cmark                       & 64.90\%                                 & 70.07\%                                   & 67.39\%           \\ \cline{3-11} 
& \textbf{\footnotesize ours} & \cmark                                  & \cmark          & \cmark                     & \cmark        & ---                          & ---                          & \textbf{70.89\%}                        & \textbf{72.11\%}                            & \textbf{71.49\%}  \\ \cline{3-11} 
        \end{tabular}
    \caption{Ablation studies of the model.} 
    \label{tab:studies}
    \vspace{-1em}
\end{table*}

\paragraph{Real Detections} We first compare how performance changes when using input crops coming from a detector as opposed to labeled bounding boxes. In particular, a lidar-based CNN is used to detect vehicles, and project the boxes into the image to crop the input for the network. Results are shown in \autoref{tab:detections}, and indicate that while performance levels are lower when using detections, the model is able to cope with imperfect ROIs and produce comparable results.

\paragraph{Attention} The impact of different attention mechanisms is considered in \autoref{tab:studies} rows $1$-$2$. Specifically, we consider a model using no attention, and a u-Net \cite{unet} based CNN for attention (marked with a $ \tightStar$).\footnote{The network consists of two, $2$-layer convolutional blocks: $32$ and $64$ channels respectively, both using kernel size $3\times3$, and no dilation.} Results show that attention increases recall by $5.48$ percentage points. The u-Net architecture did not outperform the fully convolutional approach, which can be explained by the fine positioning that the latter can achieve, emphasizing only important pixels.

\paragraph{Loss Parameterization} We show the effects of using different loss parameterizations on the model, \autoref{tab:studies} rows $3$-$4$, which affects the output produced by the model. Specifically, we consider a single task loss over the intent states, (in which case the loss function is just the $\ell_{intent}$ component), and removing supervision for the left and right sides (therefore using $\ell_{intent}$ and $\ell_{view}$ as the loss components). Results show multi-task loss helps learning, with extra supervision improving as much as $5.23$ percentage points in F$1$ between our model and its single-task counterpart.

\paragraph{Normalization} Lastly, we study the effects of adding normalization schemes on the model, \autoref{tab:studies} rows $5$-$6$. In particular, we experiment with using layer normalization (LN) and batch normalization (BN). These methods differ in terms of the dimensions in which normalization happens: layer normalization normalizes across the channel dimensions only, while batch normalization normalizes across the batch and spatial dimensions \cite{normalizers}. The results show that these methods are not beneficial for the model, which makes intuitive sense as BatchNorm is not suitable for the small batch sizes we use. One hypothesis as to why LayerNorm did not help is that, in changing the activations for each frame separately, the saliency of bright spots (turn signal lights) is diminished, which degrades performance.

\section{Conclusion}

In this paper, we have tackled the important and unexplored problem of turn signal classification. We proposed a method that can be trained end-to-end and is able to handle different viewpoints of the vehicles. The proposed network is designed to reason about both spatial and temporal features through attention, convolutions, and recurrence to classify turn signal states at the frame level for a sequence of observations. We train and evaluate our method using a dataset containing over 1.2 million  real-word images. Future works in this problem include the extension to signals from emergency vehicles and the use of more features from classification (such as images from underexposed cameras).

\section*{Acknowledgements}

The results in this publication are not an indication of the performance of Uber self-driving vehicles, as our production system utilizes a variety of algorithms and might or might not include this.

\clearpage
\bibliography{references}

\end{document}